\documentclass[conference]{IEEEtran}
\IEEEoverridecommandlockouts

\usepackage{amsmath,amssymb,amsfonts}
\usepackage{algorithmic}
\usepackage{graphicx}
\usepackage{subfig}
\usepackage{textcomp}
\usepackage{xcolor}
\usepackage{array} 
\usepackage{float}
\usepackage{hyperref}
\usepackage{cite}
\usepackage{tabularray}
\def\BibTeX{{\rm B\kern-.05em{\sc i\kern-.025em b}\kern-.08em
    T\kern-.1667em\lower.7ex\hbox{E}\kern-.125emX}}

\begin{document}

\title{Vision Controlled Orthotic Hand Exoskeleton\\}

\author{\IEEEauthorblockN{Connor Blais, Md Abdul Baset Sarker, Masudul H. Imtiaz}
\IEEEauthorblockA{\textit{Dept. of Electrical and Computer Engineering} \\
\textit{Clarkson University}\\
Potsdam, New York, USA \\
blaiscs@clarkson.edu, sarkerm@clarkson.edu, mimtiaz@clarkson.edu}

}

\maketitle

\begin{abstract}
\textbf{This paper presents the design and implementation of an AI vision-controlled orthotic hand exoskeleton to enhance rehabilitation and assistive functionality for individuals with hand mobility impairments. The system leverages a Google Coral Dev Board Micro with an Edge TPU to enable real-time object detection using a customized MobileNet\_V2 model trained on a six-class dataset. The exoskeleton autonomously detects objects, estimates proximity, and triggers pneumatic actuation for grasp-and-release tasks, eliminating the need for user-specific calibration needed in traditional EMG-based systems. The design prioritizes compactness, featuring an internal battery. It achieves an 8-hour runtime with a 1300 mAh battery. Experimental results demonstrate a 51ms inference speed, a significant improvement over prior iterations, though challenges persist in model robustness under varying lighting conditions and object orientations. While the most recent YOLO model (YOLOv11) showed potential with 15.4 FPS performance, quantization issues hindered deployment. The prototype underscores the viability of vision-controlled exoskeletons for real-world assistive applications, balancing portability, efficiency, and real-time responsiveness, while highlighting future directions for model optimization and hardware miniaturization.} 
\end{abstract}


\section{Introduction}
Orthotic hand exoskeletons are used primarily in post-stroke rehabilitation. They help minimize motor impairments through repetitive task training \cite{heo2012exoskeleton}. When used in robotic rehabilitation systems, they can provide progress tracking through embedded sensors, allowing objective assessments of rehabilitation progress. Furthermore, continuous passive motion devices automate therapeutic exercises, reducing the therapist's workload while maintaining joint mobility. However, another use case for hand exoskeletons is to enhance hand function. These can be used by individuals suffering from permanent loss of hand function or by healthy individuals. Spinal cord injuries or neuromuscular diseases are the most common causes of loss of hand function that augmentative hand exoskeletons address \cite{heo2012exoskeleton}. The exoskeleton enhances grip strength and assists in articulating their fingers to perform basic everyday tasks. The most common way to control augmentative hand exoskeletons is by using EMG sensors or an external IMU mounted elsewhere on the body \cite{heo2012exoskeleton}. \par

Industrial applications of hand exoskeletons also exist. They can be used to mitigate work-related musculoskeletal disorders caused by repetitive or forceful gripping. In addition, they can reduce muscle fatigue in assembly line workers and construction personnel. Furthermore, grip-assist systems enable prolonged package handling in the shipping industry without strain injuries. Field studies demonstrate a reduced incidence of carpal tunnel syndrome and tendinitis among users. Such preventive applications are gaining traction in industries with high repetitive stress injury rates. \par

More advanced exoskeleton platforms can also study human motor control and monitor rehabilitation goals. Such research tools provide insight into sensorimotor integration and motor learning processes. Experimental setups incorporating virtual reality interfaces examine neural adaptation during rehabilitation. These experimental setups can also help develop advanced prosthetic limbs with natural control schemes. Hybrid systems that combine functional electrical stimulation with robotic assistance explore synergistic rehabilitation approaches. These research applications contribute to evidence-based therapy protocols and standards for human-robot interaction. The data collected from these studies continuously refine exoskeleton designs for clinical and assistive use. \par
\subsection{Literature Review}
Various advancements in hand exoskeleton technologies have been made over the years. Table \ref{comptable}\footnote{A more comprehensive table can be found in \cite{heo2012exoskeleton}} summarizes three rehabilitative hands and three assistive hands that the research community has developed. 
\begin{table}
    \centering
    \begin{tblr}{@{}|X[c,h,valign=b]|X[c,h,valign=b]|X[c,h,valign=b]|X[c,h,valign=b]|X[c,h,valign=b]|@{}} \hline 
        Name & Force Transmission & Method of Intention Sensing & Type & Drive Method \\ \hline
        HandSOME \cite{Brokaw2011} & Linkage & N/A (Passive) & Rehabilitative & Passive \\ \hline
        WaveFlex \cite{OttoBock} & Linkage & N/A (CPM) & Rehabilitative & DC Motor \\ \hline
        Wege et al. \cite{Wege2007} & Cable & EMG & Rehabilitative & DC Motor \\ \hline
        Kadowaki et al. \cite{Kadowaki2011} & Glove & Flexion angle or EMG & Assistive & Pneumatic \\ \hline
        SMA Actuated Hand. \cite{patricia2018sma} & Cable Wrapped Glove & Manual Via MATLAB & Assistive & SMA \\ \hline
        AI Vision Controlled \cite{sarker2024vision} & Glove & Camera and Sensors & Assistive & Pneumatic \\ \hline
    \end{tblr}
    \smallskip
    \caption{Various Hand Exoskeleton Technologies}
    \label{comptable}
\end{table}
Among these hand exoskeletons are several different methods of driving the hand. Electric motor-driven hand exoskeletons use DC motors to generate precise, programmable forces \cite{heo2012exoskeleton}. This type of exoskeleton typically relies on cabling or a linkage structure to move the joints of the hand. Pneumatic-driven hand exoskeletons utilize compressed air to power soft actuators. Pneumatic muscles excel at applying compliant, naturalistic forces, which are beneficial for safe human-robot interaction. SMA actuation-based hand exoskeletons utilize metal alloys that contract when an electric current is applied. They are incredibly lightweight and compact because they only rely on an electrically conductive alloy to move \cite{patricia2018sma}. This allows them to be extremely low-profile, minimizing the obstruction of hand movements. Passive hand exoskeletons use springs or other elastic elements to provide a constant force to the hand. They are considered passive because they do not use any electrical components. Passive hand exoskeletons are relatively straightforward, which makes them reliable and cost-effective. However, because they have no controllable components, there is no way to adjust the force applied to the hand.\par
The last item on Table \ref{comptable} is the initial concept for the design laid out in this paper. Our research group at Clarkson University previously developed it and paired a commercially available pneumatic glove with an AI vision-based control scheme\cite{sarker2024ai}. It aimed to eliminate the need for individualized training to use an EMG-based system. The exoskeleton uses a Google Coral Dev Board Mini with an external 5-megapixel camera to detect objects in the area in front of the hand. It also uses a VL6180X time-of-flight sensor to measure the distance of objects to the palm. When an object is detected, the hand will open so that the user can grasp it. When the object is close enough, the VL6180X embedded in the wrist of the glove will trigger the glove to repressurize the pneumatics. An ADXL345 accelerometer performs gesture detection, which triggers the hand to release its holding object when the user quickly rotates their wrist \cite{sarker2024ai}. This design used an EfficientDet-Lite0 object detection model trained on a locally collected dataset. The model operated at six frames per second on the Dev Board Mini, which was deemed acceptable for real-time detection \cite{sarker2024ai}. Furthermore, a second microcontroller was also used to interface with the pump and solenoid, which controlled airflow into the hand. \par
The Google Dev Board Mini's central MCU is the MediaTek 8167s Quad-Core Arm Cortex-A35 MCU, which requires an average of $3$W to operate. Furthermore, each core of this chip can run at up to $1.5$GHz \cite{coral2025devboardmini_ds}. It also has Google's Edge Tensor Processing Unit (TPU) embedded in it. This TPU was designed to accelerate Machine Learning algorithms efficiently, which enhances the usability of Machine Learning in embedded applications \cite{coral2025devboardmini_ds}. \par
 However, this initial prototype required a system redesign and a robust object detection and gesture recognition model. Furthermore, a reduction in the size of the system was required to make it lightweight and to increase the comfort of use.

\subsection{Initial Work}
The initial design of the hand exoskeleton laid out in this paper aimed to simplify and streamline the original design. For the redesign, there were three candidates for the central microcontroller for the device. The first candidate was the Arduino Pro Portentia H7. This microcontroller was based around the STM32H747XI Dual Arm® Cortex® M7/M4 MCU; the M7 runs at 480MHz, and the M4 runs at 240MHz. In addition, it has an embedded Chrome-ART graphical hardware accelerator, which would increase the efficiency of the object detection model. The second candidate was the Arduino Pro Nicla Vision. This microcontroller has an STM32H747AII6 Dual Arm® Cortex® M7/M4 MCU; the M7 and M4 cores run at the same speed as their counterparts on the Portentia H7. Additionally, this board comes pre-equipped with a camera, IMU, and TOF sensor, and the entire board is the size of an American quarter. The third candidate was the Google Coral Dev Board Micro.  This controller has an NXP i.MX RT1176 Cortex® M7/M4 MCU; the M7 runs at 800MHz and the M4 runs at 400MHz. It also has a Coral Edge TPU coprocessor for accelerating TensorFlow Lite models, making it the fastest of the three options.\par
\begin{figure}
    \centering
    \includegraphics[width=0.7\linewidth]{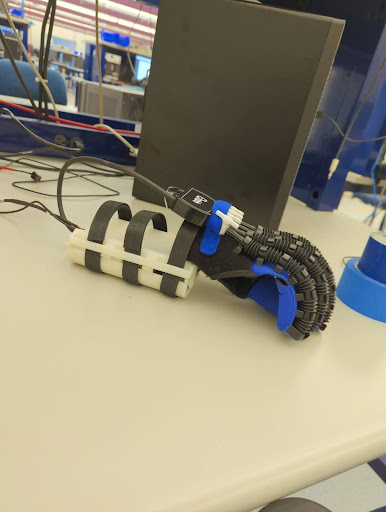}
    \caption{An overview of the proposed system}
    \label{srdesignglove}
\end{figure}

After analysis, the Nicla Vision was chosen for prototyping due to its integrated sensors and small package size. A functional standalone version (NV Version) was developed, and objects were successfully detected using a model trained with EdgeImpulse. Furthermore, it matched the performance of the previous version, running at six frames per second. However, the Nicla Vision would overheat with prolonged use and required a cooling fan to run for extended periods. The next phase of this design involved transitioning to the Portentia H7. Still, compatibility issues arose due to flawed MicroPython I$^{2}$C camera control code. Despite non-overlapping sensor addresses, the Portentia H7 could not simultaneously communicate with the camera and sensors without errors. The final design combined the Nicla Vision for object detection and the Portentia H7 for central control, sacrificing the single-microcontroller goal but achieving a compact form. This design can be seen in Figure \ref{srdesignglove}.

\subsection{Proposed Solution}
The design requirements for the revised AI vision-controlled orthotic hand exoskeleton are as follows. First, it will need to run for six to eight hours or more, as this is generally what clinicians agree is reasonable \cite{boser2021exoskeleton}. Second, it must fit entirely on the lower wrist; the battery pack should be internal, not external, as in the initial design. This makes the system easier to put on and can be contained in as small an area as possible. Third, the hand should be able to close and open. The design will use the same pneumatic glove as the earlier versions. The design will use the Google Coral Dev Board Micro as its singular microcontroller. Furthermore, the object detection model should consist of multiple classes and run at a rate of more than six frames per second. Finally, the design must be as light as possible, with the goal of less than 200 grams, as this is the maximum acceptable weight of a hand exoskeleton \cite{boser2021exoskeleton}.

\section{Theory}
For the design laid out in this paper, two different object detection model architectures were investigated, MobileNet\_V2 and YOLOv11. Furthermore, these models should run on the Google Edge TPU embedded in the Coral Micro, so their performance must be examined. The performance of the Coral Micro itself also needs to be addressed. Additionally, the embedded codebase for the Coral Micro has two relevant parts that need to be investigated: FreeRTOS and TFLiteMicro. Finally, the electrical design for the custom motor controller needs to be explored.

\subsection{AI Vision Control}
MobileNet\_V2 is a deep learning model architecture developed by Google for use in mobile environments \cite{sandler2019mobilenetv2}. It is based on MobileNetV1 and uses Linear Bottlenecks to enhance the model's efficiency and maintain a high accuracy. These Linear Bottlenecks help prevent information loss caused by non-linear activation functions, like ReLU. Because the bottlenecks in the model are linear, more critical information is preserved, allowing for better expressiveness \cite{sandler2019mobilenetv2}. Another significant change MobileNet\_V2 makes is using Inverted Residual Blocks, which place shortcuts between the bottleneck layers instead of between broad feature representations like typical Residual Blocks would do. This allows for the efficient reuse of features at lower computational costs. Furthermore, MobileNet\_V2 makes heavy use of depthwise separable convolutions, which reduce the number of computations needed compared to traditional convolutions. These improvements from MobileNetV1 allow MobileNet\_V2 to approach the performance of large models while using less memory and still having a low latency \cite{sandler2019mobilenetv2}.\par 

YOLOv11, or You Only Look Once v11, is the latest version of YOLO that has improved upon its predecessors, emphasizing its ability to detect objects in real-time. Additionally, it has enhanced capabilities to do instance segmentation and pose estimation compared to its previous versions. YOLO is known for its fast and accurate classification abilities; every iteration continuously improves upon this. YOLOv11's most significant improvements from its predecessors are using C3k2 Blocks for feature extraction, SPPF modules for spatial pyramid pooling, and C2PSA blocks for spatial attention. These changes give YOLOv11 higher accuracy and faster interpretation speeds than its predecessor, making it a good choice for low-latency scenarios \cite{khanam2024yolov11}. \par

\subsection{Edge TPU}
The Edge TPU developed by Google is an ML Accelerator with significant speed advantages in embedded applications due to its specialized architecture \cite{seshadri2022edgetpu}. It achieves remarkably low inference latency, which is critical for real-time edge AI tasks. It can run MobileNet\_V2 models at speeds as high as 400 FPS \cite{coral2025accelerator}. High-accuracy models (e.g., $95\%$ validation accuracy) run in as little as 4.18 ms on the V2 configuration, while smaller models with optimized operations, such as 1×1 convolutions, achieve sub-millisecond latencies (e.g., 0.074 ms). Performance varies by workload: V1 excels for large models (5–30M parameters) due to its 2 MB PE memory, minimizing off-chip data transfers, whereas V3 leverages architectural tweaks (e.g., more cores per PE) to deliver 10.4× speedups over V1 for models dominated by 1×1 convolutions \cite{seshadri2022edgetpu}. These optimizations ensure rapid inference across diverse neural network structures.\par
The Edge TPU is also very energy efficient, using only half a watt per TOPS. Employing parameter caching and reusing on-chip weights across inferences reduces costly off-chip memory accesses. For instance, V2 consumes 19.75 mJ for high-accuracy models, slightly outperforming V1 (19.89 mJ) in similar tasks. Smaller models with fewer than 3M parameters have an energy consumption of 0.17 mJ on the V2. This makes the Edge TPU suitable for battery-operated embedded applications \cite{seshadri2022edgetpu}. \par
The accelerator's hardware parallelism further boosts performance. Each processing element (PE) integrates SIMD lanes and multi-core designs to execute convolutional operations in parallel. For example, a PE with four SIMD lanes processes four kernel elements simultaneously, accelerating loop-heavy computations like convolutions. Additionally, the Edge TPU's software ecosystem supports rapid design exploration. A graph neural network (GNN) based learned model predicts millisecond latency and energy, bypassing hours-long cycle-accurate simulations. \par
\subsection{Coral Dev Board Micro}
The Coral Dev Board Micro is designed as a high-performance embedded system centered around Google's Edge TPU, which delivers 4 TOPS (trillion operations per second) for accelerated inference of quantized TensorFlow Lite models \cite{imran2020embedded}. This capability enables real-time processing of machine learning tasks, such as image classification or audio analysis, directly on the edge without cloud dependency. The Edge TPU interfaces with the dual-core NXP i.MX RT1176 MCU (800 MHz Cortex-M7 and 400 MHz Cortex-M4) via USB 2.0, creating a balance between preprocessing on the ARM cores and compute-heavy inferencing on the Edge TPU. While the USB 2.0 link limits data bandwidth to 480 Mbps, the system optimizes efficiency by directly integrating a 324x324-pixel camera on the board, minimizing external data transfers and enabling streamlined sensor-to-inference pipelines. This architecture uses high-speed processing for real-time vision or voice recognition applications. Furthermore, the NXP i.MX RT1176 MCU, the board's central processor, is very low power, drawing less than $1W$ during regular operation \cite{coral2025devboardmicro_ds}.

\subsection{Embedded Codebase}
\begin{figure}
    \centering
    \includegraphics[width=1\linewidth]{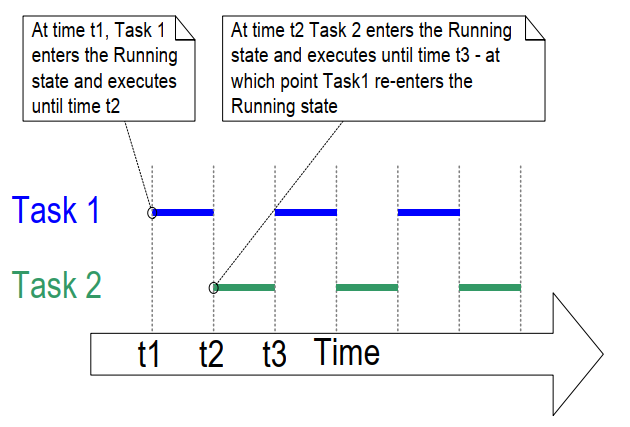}
    \caption{FreeRTOS Task Execution \cite{freertos2025kernel}}
    \label{rtostasks}
\end{figure}

FreeRTOS is a lightweight, open-source, real-time operating system (RTOS) designed for embedded systems with deterministic behavior and minimal resource overhead. At its core, FreeRTOS employs a preemptive, priority-based scheduler to manage tasks (threads). Each task is assigned a priority level, and the scheduler ensures the highest-priority-ready task always runs first. For tasks of equal priority, FreeRTOS supports round-robin scheduling, where tasks share CPU time in fixed time slices; see Figure \ref{rtostasks}. The kernel maintains task states (ready, running, blocked, suspended) and uses a doubly linked list to manage tasks at the same priority level \cite{amos2020rtos}. The pxCurrentTCB pointer is a key part of FreeRTOS. It tracks the currently running task in conjunction with the pxReadyTasksLists array, which organizes ready tasks by priority. Context switching occurs during scheduler ticks, task yields, or synchronization events, with the scheduler saving and restoring task contexts (registers, stack) to ensure seamless transitions. \par
TensorFlow Lite Micro (TFLiteMicro) is a specialized variant of TensorFlow Lite (TFLite) designed to run machine learning models on microcontrollers and deeply embedded systems with extreme resource constraints, such as kilobytes of memory, no operating system, and no dynamic memory allocation. While TFLite targets mobile and edge devices with moderate resources, leveraging an interpreter to execute models and supporting features like dynamic memory and hardware acceleration, TFLite Micro strips away these assumptions to operate in environments where even basic OS features are absent \cite{warden2019tinyml}. TFLite Micro requires applications to pre-allocate a static memory arena. This memory arena is managed with two stacks, one for persistent data and another for transient buffers. This eliminates heap fragmentation, ensuring predictable memory usage. During initialization, the interpreter plans memory layouts using bin-packing algorithms to maximize reuse, and all allocations are finalized upfront, avoiding runtime memory management overhead. TFLite Micro retains compatibility with most of TensorFlow's model conversion pipeline and quantization tools. This compatibility ensures that TFLite Micro can adapt to a wide range of TensorFlow models, making it ideal for low-power applications that require continuous operation. Furthermore, it can still be used for resource-intensive tasks, such as object detection, provided that the model does not rely on operations not supported by TFLite Micro.
\subsection{Electrical Design}
Two $5V$ vacuum pumps will be used to control the motion of the hand, one to open the hand and one to close it. The vacuum pumps that will be used are polarized DC motors, which means that the airflow direction can not be changed by reversing the voltage across the motor. Because of this, an H-Bridge controller is unnecessary, so a MOSFET DC Motor driver can be used. To design this driver properly, the electrical characteristics of the motors need to be determined. Doing this yields values of $2mH$ and $3.99\Omega$. These values can be used to model the motor in LTSpice roughly \cite{burroughs2025spice}.
\par Additionally, the motor's current draw must be measured at the various stages of motor operation so that we can adequately account for the load the motor is experiencing in our model. Attaching the motor to the hand exoskeleton on the intake and outtake ports and running it yields a current draw of $250mA$ while opening the hand and a current draw of $395mA$ while closing the hand. From here, we can simulate the behavior of the motor in LTSpice and use this to verify that the MOSFET-based driver will not exceed its power dissipation rating.
\par The solenoid that will be used to swap between inlet and outlet is much easier to model in LTSpice, as it is just an inductor. Measuring its characteristics yields values of $6.73mH$ and $23.1\Omega$. Furthermore, connecting the solenoid to a power supply of $3.3V$ shows that it draws $130mA$ on average. These values can then be used in LTSpice to model the solenoid and check if the motor diver needs modification. Doing this shows that the driver can handle the solenoid without any modifications. Furthermore, a flyback diode is necessary across the motors and the solenoid to account for voltage spikes caused by their inductance.\par 
This design needs a power MOSFET with a low threshold voltage, reasonable power dissipation, and a low $R_{DS(on)}$. The IRLML6244TRPbF fits these requirements perfectly \cite{irlml2012}. It has a threshold voltage between $0.5V$ and $1.1V$, which gives plenty of headroom for the $1.8V$ logic level the Coral Micro operates on. It has a max power dissipation of $1.3W$. Its $R_{DS(on)}$ is between $22m\Omega$ and $27m\Omega$ at a $V_{GS}$ of $2.5V$, which will allow for minimal power loss when driving the motors and solenoids. Additionally, it comes in a SOT23-3 package, which is very small and will allow for a more compact design.
\par First, an approximate $R_{DS(on)}$ for a $V_{GS}$ of $1.8V$ needs to be calculated. The easiest method to determine this is to use LTSpice, as there is a built-in IRLML6244TRPbF model. Assuming the voltage drop across the pump gives a $V_{D}$ of no more than $100mV$, the simulation shows that $R_{DS(on)}$ is between $52m\Omega$ and $54m\Omega$. While this is much higher than the values listed on the data sheet, those values are based on a $V_{GS}$ of $4.5V$ and $2.5V$ respectively \cite{irlml2012}. Even so, $54m\Omega$ is sufficiently low for the driver to minimize power dissipation during operation.
\par Second, using LTSpice to simulate a simple MOSFET switch to drive the motor reveals a potential problem. This problem is a significant power dissipation spike when the MOSFET closes. Fortunately, this problem can be solved by adding a soft start as a low-pass filter to the MOSFET Gate. Using a $100k\Omega$ resistor and a $1\mu F$ capacitor decreases the power dissipation of the MOSFET to well below $1.3W$ when the MOSFET closes.
\section{Implimentation}
\subsection{Firmware Design}
\subsubsection{Coral Micro Codebase}
The source code for the Coral Micro is based on FreeRTOS. This design's code is split into four tasks: InfrenceTask, SensorTask, MotorTask, and BatteryTask. Each task has a priority level, with MotorTask and BatteryTask having the highest priority, InfrenceTask having the second highest priority, and SensorTask having the lowest. Flowcharts for MotorTask, BatteryTask, InfrenceTask, and SensorTask can be found in \ref{appendix} Appendix. A video of the device in operation can be found here: \href{https://youtu.be/AKla_7vtGEA}{Link}\footnote{https://youtu.be/AKla\_7vtGEA}.\par
MotorTask manages the state machine that controls the exoskeleton. It has three states: OpenHand, CloseHand, and Idle. When InfrenceTask triggers the OpenHand state, MotorTask suspends InfrenceTask, opens the hand, and records the system tick count. It then checks if the VL6180X detects an object within 30mm of the sensor. If it does not, MotorTask delays for 100ms; if it does detect something, it sets the state to CloseHand and clears the TapDetected registers of the ADXL345. If the VL6180X does not detect any object within 30mm for 10 seconds, MotorTask sets the hand to its rest state\footnote{The rest state is when the glove is at atmospheric pressure internally}, resumes the InfrenceTask, and then returns to its Idle state. When in the CloseHand state, MotorTask checks the TapDetected registers of the ADXL345 every 100ms. If the ADXL345 has detected a tap, MotorTask opens the hand and sets it to its rest state. It then resumes InfrenceTask and returns to its Idle state. In its Idle state, it waits for InfrenceTask to trigger the OpenHand state.
BatteryTask checks the voltage of the battery every minute. It suspends all tasks if it calculates a runtime of less than 30 minutes. At the start of the loop, it records the current system tick count. It then checks the battery voltage every minute. Once it has ten measurements, it records the current system tick count again. It then calculates how much time has passed since it started recording the first of the 10 measurements. If the last battery level recorded is higher than its current level, it estimates the remaining runtime of the exoskeleton. The runtime estimation is a simple linear approximation that assumes that the minimum battery voltage the device will operate at is 9V. 9V is when the power regulator's voltage drops below the minimum required to run the Coral during inference. All other tasks are suspended if the runtime is less than thirty minutes. BatteryTask will continue to check the battery level every minute until the system is powered down.\par

InfenceTask starts by initializing the TPU and the necessary software components to run the object detection model. This setup is based on a tutorial by Shawn Himel, who based it on the object detection examples provided by Google for the Coral Micro\cite{digikey2025coral}. After setup, InfrenceTask enters its main loop, where it continuously captures a frame from the camera and determines if any objects are present. If it detects an object, and the ID of the last object it detected is not zero, it turns the laser off. Then, if the current object is the same as the object detected in the previous frame, it increments the object count by one. It then checks if the object count is greater than five and that the state of MotorTask is not OpenHand\footnote{This is done so that in the event InfrenceTask gets scheduled to run multiple times before MotorTask can suspend it, InfrenceTask does not continually try to set MotorTask's state}. If both are true, it resets the object count and then turns on the ADXL345's tap detection. Then, it sets the MotorTasks state to OpenHand. If the same object is not detected for six frames in a row, the object count is reset, and the last object ID is updated to the new object detected. \par

SensorTask records data from the AXDL345 and the VL6180X. It saves this data to two global buffers so that the other tasks can use them for control purposes. In addition to this, there is a watchdog timer that triggers after 8 seconds of inactivity. This prevents the power from needing to be cycled if something goes wrong on the board.
\subsubsection{MobileNet\_V2 Training}
The MobileNet\_V2 model was trained with a dataset consisting of 1923 images, split across six classes. These six classes are: 'ball', 'bottle', 'cube', 'cup', 'pen', and 'spoon'. Before training, all the images in the dataset are scaled to 324x324, with padding added so that the aspect ratio is retained. Furthermore, to increase the variety of data, each image is rotated between -45 and 45 degrees, such that each original image becomes ten images. Applying this transformation increases the total number of images in the set to 19230. The dataset is then split into train, test, and validation with an 80-10-10 split. \par

The model training is done with the MediaPipe-Model-Maker Python library following the standard procedures provided in \cite{digikey2025coral}. It uses the MOBILENET\_V2\_I320 pretrained model as its base and retrains it with the parameters in Table \ref{paramtable}. 
\begin{table}
    \centering
    \begin{tabular}{|l|c|}
        \hline
        Parameter & Value \\
        \hline
        Learning Rate & 0.04 \\
        Batch Size & 16 \\
        Epochs & 100 \\
        Cosine Decay Alpha & 0.98 \\
        Cosine Decay Epochs & 50 \\
        \hline
    \end{tabular}
    \smallskip
    \caption{Training Parameters}
    \label{paramtable}
\end{table}

Doing this yields a model with an AP value of .67 and a loss of 0.3368. The final model is also quantized to int8 and compiled for the Edge TPU.
\section{Results}
\subsection{Model Performace}
\begin{figure}
    \centering
    \subfloat[Ball Overhead Light]{\includegraphics[width=0.45\linewidth]{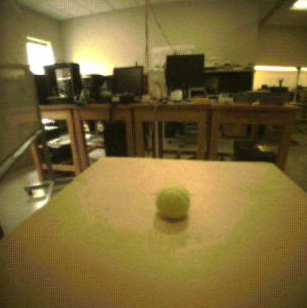}}
    \subfloat[Ball Overhead Light with Flashlight]{\includegraphics[width=0.45\linewidth]{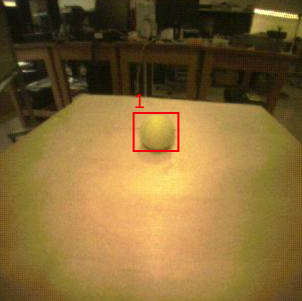}}\\
    \subfloat[Cup Overhead Light]{\includegraphics[width=0.45\linewidth]{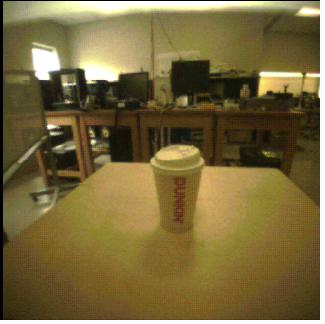}}
    \subfloat[Cup Overhead Light with Flashlight]{\includegraphics[width=0.45\linewidth]{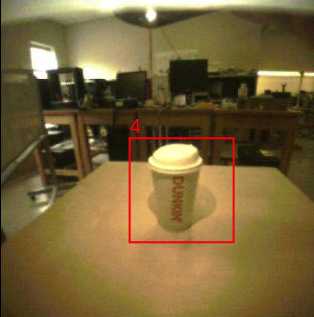}}\\
    \subfloat[Cup No Overhead Light]{\includegraphics[width=0.45\linewidth]{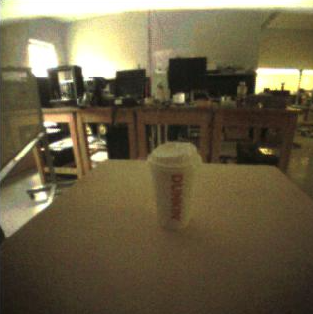}}
    \subfloat[Cup No Overhead Light with Flashlight]{\includegraphics[width=0.45\linewidth]{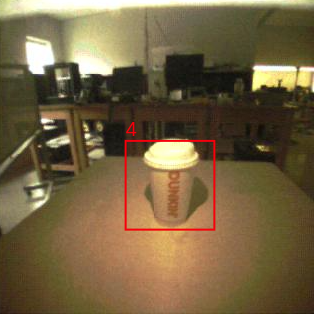}}
    \caption{Lighting Tests}
    \label{lightingtest}
\end{figure}

When implemented on the Coral Micro, the MobileNet\_V2 model runs at 10.0 FPS. This is a significant improvement from the previous versions of this design, which ran at 6.0 FPS. That being said, 10.0 FPS is a considerable improvement over typical tflite models, which run at around 3.5 FPS on mobile phones \cite{tflite2025embedded}. However, the model struggles to recognize the pen and spoon classes reliably. For the pen class specifically, the orientation of the pen in the frame determines how often it is detected. Furthermore, the bottle and cup classes overlap in their detection. This is not a huge concern, as they are similar in how they must be picked up and put down. A more considerable concern is that the model struggles to function depending on the light level of the environment and how the object is lit in frame. For example, with the cup class, with overhead lights on a bright surface, the model would only detect that the cup was there for one frame, roughly every 2 seconds. Then, the model would not detect the cup with the overhead lights off. Then, in both conditions, using a flashlight to light the cup from a roughly $45$ degree angle, the model would continuously detect the cup in frame with no issue. See Figure \ref{lightingtest} for an example of each condition. Similar results were also achieved with the ball and cube classes. Furthermore, the model is overfit to the objects within the dataset and is mostly unable to recognize objects it has never seen before.
The YOLOv11 model runs at 15.4 FPS when implemented on the Coral Micro. This is over double the frame rate of the initial design. Unfortunately, the YOLOv11 model does not correctly detect objects in the dataset. Instead, it constantly outputs the same bounding boxes with the identical object IDs. Running the model on a computer with Python gives the same result. Further investigation showed that the unquantized model performed as expected, with an AP of $99\%$, meaning there was an issue with the quantization process. The quantized model had an AP of less than $50\%$. Because of this, the YOLOv11 model will not be implemented until the quantization process can be adapted.
\subsection{Ease of Use}
When idle, the exoskeleton draws an average of 100mA. When opening, it draws an average of 250mA. When closing, it draws 625mA. When holding an object, it draws an average of 230mA. Running the hand in its idle state with an object for it to detect, with an average of two detection cycles per minute, causes the battery voltage to drop from $12.8V$ to $11.19V$ over 8 hours for a $1300$ mAh, $12$V battery. Running the hand in its hold state exclusively causes the same battery voltage to drop from $12.89V$ to $11.35V$ over 3 hours. The cut-off voltage for the system to hibernate is $9V$, so it is safe to say that the system has an 8+ hour runtime during regular operation if a $1300$mAh $12$V battery is used. $9V$ was chosen as the cut-off voltage because the $5V$ regulator starts to dip well below $5V$ when drawing more than $1A$ from it.\par
From a performance standpoint, the hand performs moderately well. The code works as intended and does not crash, and the hardware does not fail significantly. The object detection model underperforms, as getting it to recognize objects consistently can be tedious. More often than not, you need to find the sweet spot with how the camera is angled at the object. This is partly due to how the code is structured, as it requires six frames in a row to contain the same object before the hand will open. This was done to mitigate the chances of a false detection event. Theoretically, this should not cause issues as the model's validation accuracy is 67\% 
Furthermore, the hand needs to be held still more often than not to detect an object, which can pose problems for individuals suffering from neurological diseases using an orthotic hand exoskeleton.
\section{Discussion}
Several technical challenges were encountered during development, primarily due to software compatibility and hardware limitations. Setting up CUDA 12.1 on Ubuntu 18.04 required manual installation due to lab constraints and MediaPipe Model Maker's dependency on newer CUDA versions, which conflicted with the OS's default installer. Further complications arose when MediaPipe automatically installed TensorFlow-CPU, necessitating a reinstallation of TensorFlow-GPU. YOLOv11 posed significant hurdles: training crashes linked to the faulty quantization code, which were later resolved. Furthermore, compiling the model for Edge TPU required reducing input resolution to 160x160 to prevent crashes. On the Coral Micro, quantized YOLOv11 models failed to run correctly. When run, it would detect the same objects with similar scores in the same place in the frame, regardless of what was present in the frame. Hardware challenges, I$^{2}$C sensor source code incompatibility, and reliance on a dev branch of the Coral Micro source code to get TFLite Micro to support critical operations. Furthermore, the team only tested this prototype in a controlled lab environment. More rigorous testing in real-world environments is needed before this solution can be implemented on a large scale.
\section{Future Work}
Several parts of this design could be improved upon in future works. The first thing that could be improved upon is the compactness of the design. The design is taller than the previous version because of how the Coral Micro is mounted. Furthermore, the additional pump caused the design to be wider than the earlier version. The mounting of the Coral Micro could be changed by embedding it in a custom PCB or making a custom camera connector. Both options would allow the board to be mounted flat on the wrist instead of perpendicular to it. Smaller pumps could be used to decrease the width and height of the design further. Another area of improvement would be to have individual control of each finger. This would allow for different grab patterns depending on what object is detected. Another area to investigate would be using a lightweight machine learning model to detect when the user wants to put down an object. Switching over to a 6-axis IMU would be beneficial for this. Another thing that could be done is to use a stronger pump for hand closing so that the grip strength of the device is higher. To this end, a grippy material could also be added to the fingertips. Finally, the most significant area for improvement is the object detection model. More augmentations, like zoom and light level adjustment, need to be used to increase the model's accuracy in different conditions. Furthermore, each dataset class must incorporate a more diverse set of objects.
\section{Conclusion}
This AI Vision Controlled Orthotic Hand Exoskeleton demonstrates how lightweight, intelligent systems can enhance rehabilitation and assistive technology. By leveraging the Coral Dev Board Micro and an optimized MobileNet\_V2 model, the device achieves real-time object detection at 10 FPS while maintaining a compact, wearable design. The electrical and mechanical improvements, such as efficient power management allowing for a runtime of over 8 hours and a comfortable forearm-mounted case, ensure practicality for daily use. \par
Challenges do remain and leave room for future improvements. The MobileNet\_V2 model, while functional, does struggle with particular objects and lighting conditions. The YOLOv11 model, while compatible on paper, does not have the proper quantization methods necessary to make accurate inferences. Beyond its technical performance, this project underscores the real-world potential of AI-powered exoskeletons: eliminating the need for calibration for specific users moves us closer to a seamless assistive experience. Ultimately, this exoskeleton is a step toward more responsive, user-driven assistive devices that empower individuals with limited hand mobility.

\newpage
\pagenumbering{Roman}

\onecolumn
\section{Appendix}\label{appendix}
\begin{figure}[H]
    \centering
    \includegraphics[width=.9\linewidth]{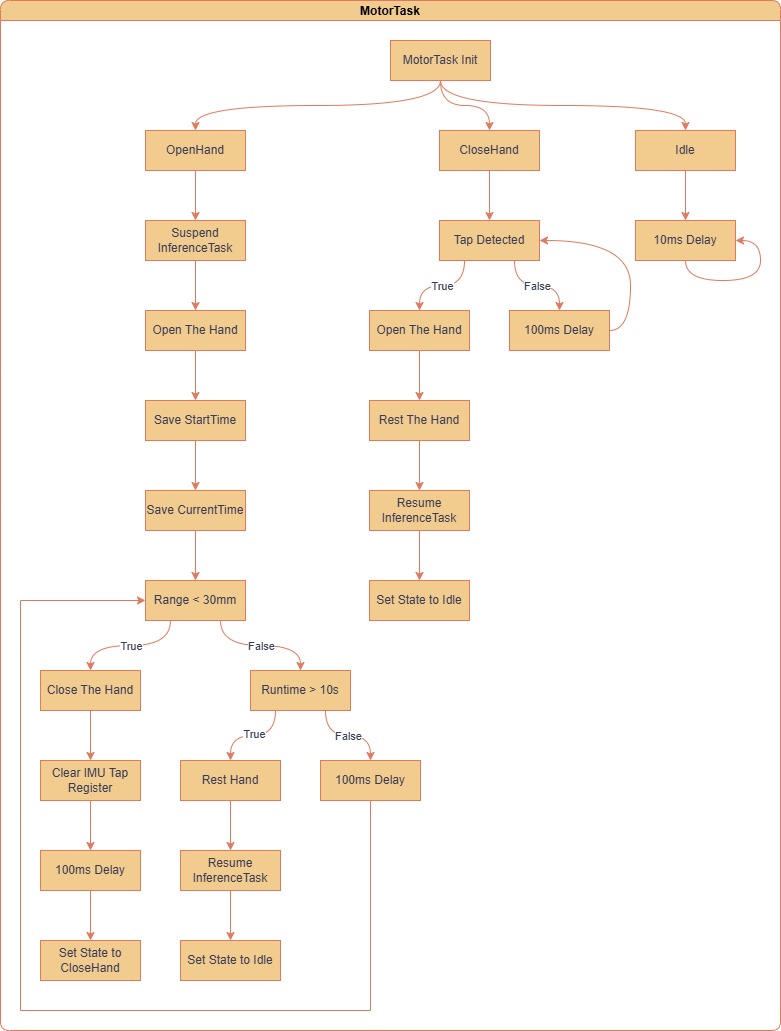}
    \caption{Flowchart of MotorTask}
    \label{MotorTaskFlow}
\end{figure}
\begin{figure}
    \centering
    \includegraphics[width=.45\linewidth]{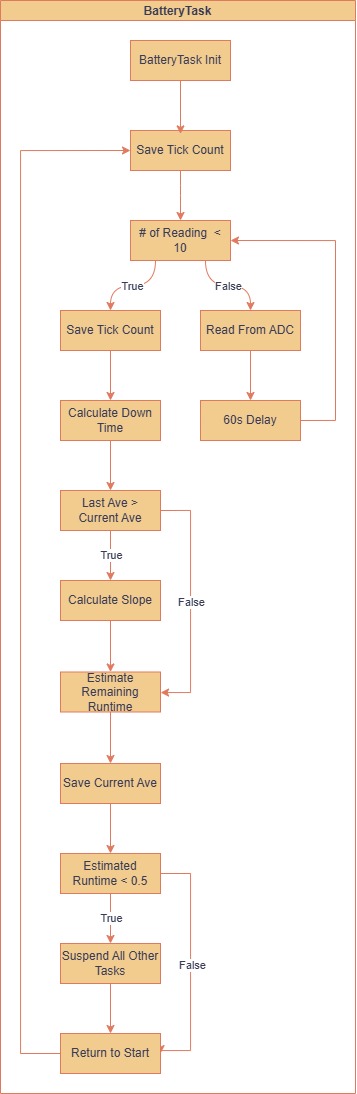}
    \caption{Flowchart of BatteryTask}
    \label{BatteryTaskFlow}
\end{figure}
\begin{figure}
    \centering
    \includegraphics[width=0.35\linewidth]{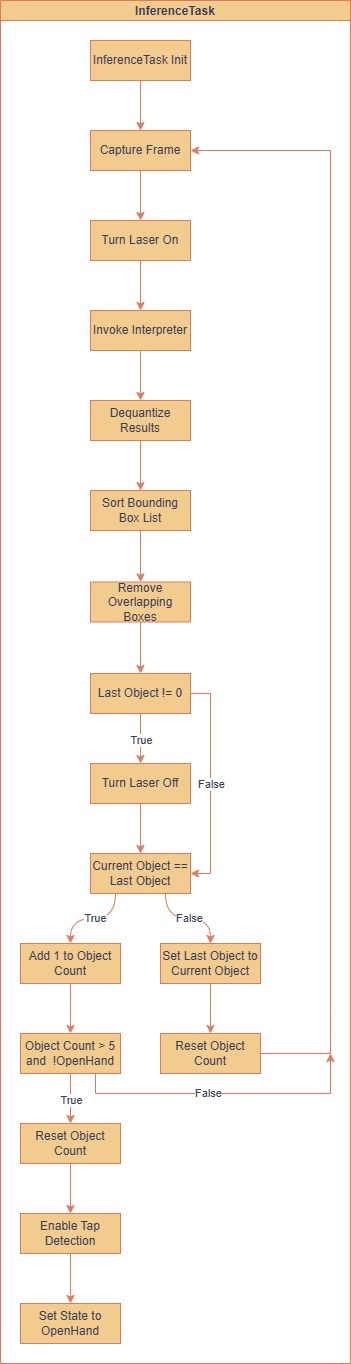}
    \caption{Flowchart of InferenceTask}
    \label{InferenceTaskFlow}
\end{figure}
\begin{figure}
    \centering
    \includegraphics[width=0.5\linewidth]{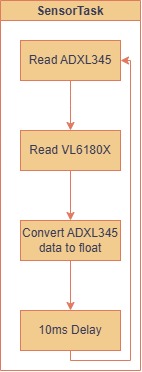}
    \caption{Flowchart of SensorTask}
    \label{SensorTaskFlow}
\end{figure}

\end{document}